# Quantum Computing and Phase Transitions in Combinatorial Search

**Tad Hogg**                                                                HOGG@PARC.XEROX.COM
*Xerox Palo Alto Research Center*
*3333 Coyote Hill Road Palo Alto, CA 94304 USA*

## Abstract

We introduce an algorithm for combinatorial search on quantum computers that is capable of significantly concentrating amplitude into solutions for some NP search problems, on average. This is done by exploiting the same aspects of problem structure as used by classical backtrack methods to avoid unproductive search choices. This quantum algorithm is much more likely to find solutions than the simple direct use of quantum parallelism. Furthermore, empirical evaluation on small problems shows this quantum algorithm displays the same phase transition behavior, and at the same location, as seen in many previously studied classical search methods. Specifically, difficult problem instances are concentrated near the abrupt change from underconstrained to overconstrained problems.

## 1. Introduction

Computation is ultimately a physical process (Landauer, 1991). That is, in practice the range of physically realizable devices determines what is computable and the resources, such as computer time, required to solve a given problem. Computing machines can exploit a variety of physical processes and structures to provide distinct trade-offs in resource requirements. An example is the development of parallel computers with their trade-off of overall computation time against the number of processors employed. Effective use of this trade-off can require algorithms that would be very inefficient if implemented serially.

Another example is given by hypothetical quantum computers (DiVincenzo, 1995). They offer the potential of exploiting quantum parallelism to trade computation time against the use of coherent interference among very many different computational paths. However, restrictions on physically realizable operations make this trade-off difficult to exploit for search problems, resulting in algorithms essentially equivalent to the inefficient method of generate-and-test. Fortunately, recent work on factoring (Shor, 1994) shows that better algorithms are possible. Here we continue this line of work by introducing a new quantum algorithm for some particularly difficult combinatorial search problems. While this algorithm represents a substantial improvement for quantum computers, it is particularly inefficient as a classical search method, both in memory and time requirements.

When evaluating algorithms, computational complexity theory usually focuses on the scaling behavior in the worst case. Of particular theoretical concern is whether the search cost grows exponentially or polynomially. However, in many practical situations, typical or average behavior is of more interest. This is especially true because many instances of search problems are much easier to solve than is suggested by worst case analyses. In fact, recent studies have revealed an important regularity in the class of search problems. Specifically, for a wide variety of search methods, the hard instances are not only rare but also concentrated near abrupt transitions in problem behavior analogous to physical phase transitions (Hogg, Huberman, & Williams, 1996). To exhibit this concentration of hard instances a search algorithm must exploit the problem constraints to prune unproductive search choices. Unfortunately, this is not easy to do within the range of allowable quantum computational operations. It is thus of interest to see if these results generalize to quantum search methods as well.

In this paper, the new algorithm is evaluated empirically to determine its average behavior. The algorithm is also shown to exhibit the phase transition, indicating it is indeed managing to, in effect, prune unproductive search. This leaves for future work the analysis of its worst case performance.

This paper is organized as follows. First we discuss combinatorial search problems and the phase transitions where hard problem instances are concentrated. Second, after a brief summary of quantum computing, the new quantum search algorithm is motivated and described. In fact, there are a number of natural variants of the general algorithm. Two of these are evaluated empirically to exhibit the generality of the phase transition and their performance. Finally, some important caveats for the implementation of quantum computers and open issues are presented.

## 2. Combinatorial Search

Combinatorial search is among the hardest of common computational problems: the solution time can grow exponentially with the size of the problem (Garey & Johnson, 1979). Examples arise in scheduling, planning, circuit layout and machine vision, to name a few areas. Many of these examples can be viewed as constraint satisfaction problems (CSPs) (Mackworth, 1992). Here we are given a set of $n$ variables each of which can be assigned $b$ possible values. The problem is to find an assignment for each variable that together satisfy some specified constraints. For instance, consider the small scheduling problem of selecting one of two periods in which to teach each of two classes that are taught by the same person. We can regard each class as a variable and its time slot as its value, i.e., here $n = b = 2$. The constraints are that the two classes are not assigned to be at the same time.

Fundamentally, the combinatorial search problem consists of finding those combinations of a discrete set of items that satisfy specified requirements. The number of possible combinations to consider grows very rapidly (e.g., exponentially or factorially) with the number of items, leading to potentially lengthy solution times and severely limiting the feasible size of such problems. For example, the number of possible assignments in a constraint problem is $b^n$, which grows exponentially with the problem size (given by the number of variables $n$).

Because of the exponentially large number of possibilities it appears the time required to solve such problems must grow exponentially, in the worst case. However for many such problems it is easy to verify a solution is in fact correct. These problems form the well-studied class of NP problems: informally we say they are hard to solve but easy to check. One well-studied instance is graph coloring, where the variables represent nodes in a graph, the values are colors for the nodes and the constraints are that each pair of nodes linked by an edge in the graph must have different colors. Another example is propositional satisfiability (SAT), where the variables take on logical values of true or false, and the assignment must satisfy a specified propositional formula involving the variables. Both these examples are instances of particularly difficult NP problems known as the class of NP-complete search problems (Garey & Johnson, 1979).

## 2.1 Phase Transitions

Much of the theoretical work on NP search problems examines their worst case behavior. Although these search problems can be very hard, in the worst case, there is a great deal of individual variation in these problems and among different search methods. A number of recent studies of NP search problems have focused on regularities of the typical behavior (Cheeseman, Kanefsky, & Taylor, 1991; Mitchell, Selman, & Levesque, 1992; Williams & Hogg, 1994; Hogg et al., 1996; Hogg, 1994). This work has identified a number of common behaviors. Specifically, for large problems, a few parameters characterizing their structure determine the relative difficulty for a wide variety of common search methods, on average. Moreover, changes in these parameters give rise to transitions, becoming more abrupt for larger problems, that are analogous to phase transitions in physical systems. In this case, the transition is from underconstrained to overconstrained problems, with the hardest cases concentrated in the transition region. One powerful result of this work is that this concentration of hard cases occurs at the same parameter values for a wide range of search methods. That is, this behavior is a property of the problems rather than of the details of the search algorithm.

This can be understood by viewing a search as making a series of choices until a solution is found. The overall search will usually be relatively easy (i.e., require few steps) if either there are many choices leading to solutions or else choices that do not lead to solutions can be recognized quickly as such, so that unproductive search is avoided. Whether this condition holds is in turn determined by how tightly constrained the problem is. When there are few constraints almost all choices are good ones, leading quickly to a solution. With many constraints, on the other hand, there are few good choices but the bad ones can be recognized very quickly as violating some constraints so that not much time is wasted considering them. In between these two cases are the hard problems: enough constraints so good choices are rare but few enough that bad choices are usually recognized only with a lot of additional search.

A more detailed analysis suggests a series of transitions (Hogg & Williams, 1994). With very few constraints, the average search cost scales polynomially. As more constraints are added, there is a transition to exponential scaling. The rate of growth of this exponential increases until the transition region described above is reached. Beyond that point, with its concentration of hard problems, the growth rate decreases. Eventually, for very highly constrained problems, the search cost again grows only polynomially with size.

## 2.2 The Combinatorial Search Space

A general view of the combinatorial search problem is that it consists of $N$ items[1] and a requirement to find a solution, i.e., a set of $L < N$ items that satisfies specified conditions or constraints. These conditions in turn can be described as a collection of *nogoods*, i.e., sets of items whose combination is inconsistent with the given conditions. In this context we define a *good* to be a set of items that is consistent with all the constraints of the problem. We also say a set is *complete* if it has $L$ items, while smaller sets are *partial* or *incomplete*. Thus a solution is a complete good set. In addition, a *partial solution* is an incomplete good set.

---

1. For CSPs, these items are all possible variable-value pairs.

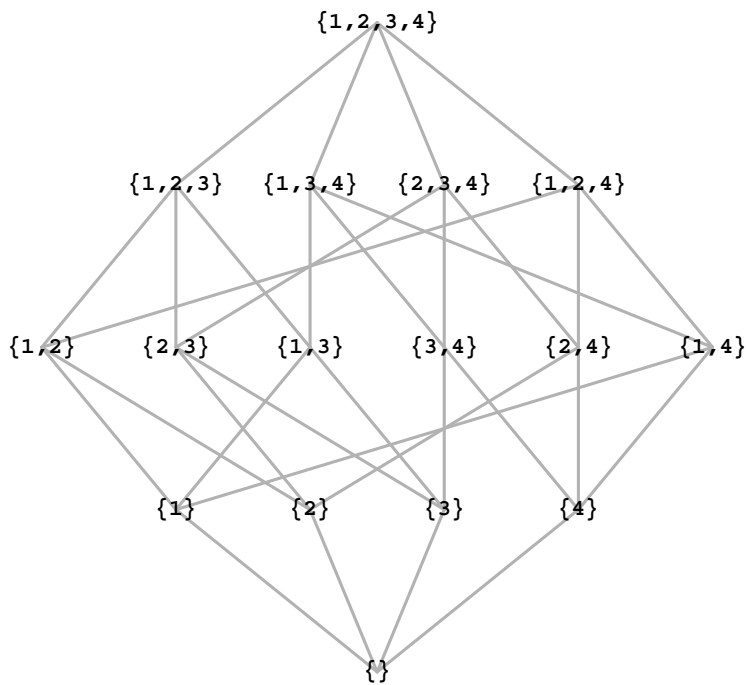

Figure 1: Structure of the set lattice for a problem with four items. The subsets of $\{1, 2, 3, 4\}$ are grouped into levels by size and lines drawn between each set and its immediate supersets and subsets. The bottom of the lattice, level 0, represents the single set of size zero, the four points at level 1 represent the four singleton subsets, etc.

A key property that makes this set representation conceptually useful is that if a set is nogood, so are all of its supersets. These sets, grouped by size and with each set linked to its immediate supersets and subsets, form a lattice structure. This structure for $N = 4$ is shown in Fig. 1. We say that the

$$N_i = \binom{N}{i} \tag{1}$$

sets of size $i$ are at level $i$ in the lattice. As described below, the various paths through the lattice from levels near the bottom up to solutions, at level $L$, can be used to create quantum interference as the basis for a search algorithm.

As an example, consider a problem with $N = 4$ and $L = 2$, and suppose the constraints eliminate items 1 and 3. Then we have the sets $\{\}$, $\{2\}$, and $\{4\}$ as partial goods, while $\{1\}$ and $\{3\}$ are partial nogoods. Among the 6 complete sets, only $\{2,4\}$ is good as the others are supersets of $\{1\}$ or $\{3\}$ and hence nogood.

For the search problems studied here, the nogoods directly specified by the problem constraints will be small sets of items, e.g., of size two or three. On the other hand, the number of items and the size of the solutions will grow with the problem size. This gives a number of small nogoods, i.e., near the bottom of the lattice. Examples of such problems include binary constraint satisfaction, graph coloring and propositional satisfiability mentioned above.

For CSPs, the items are just the possible variable-value pairs in the problem. Thus a CSP with $n$ variables and $b$ values for each has $N = nb$ items[2]. A solution consists of an assignment to each variable that satisfies whatever constraints are given in the problem. Thus a solution consists of a set of $L = n$ items. In terms of the general framework for combinatorial search these constraint satisfaction problems will also contain a number of problem-independent *necessary* nogoods, namely those corresponding to giving the same variable two different values. There are $n \binom{b}{2}$ such necessary nogoods. For a nontrivial search we must have $b \geq 2$, so we restrict our attention to the case where $L \leq N/2$. This requirement is important in allowing the construction of the quantum search method described below.

Another example is given by a simple CSP consisting of $n = 2$ variables ($v_1$ and $v_2$) each of which can take on one of $b = 2$ values (1 or 2) and the single constraint that the two variables take on distinct values, i.e., $v_1 \neq v_2$. Hence there are $N = nb = 4$ variable-value pairs $v_1 = 1$, $v_1 = 2$, $v_2 = 1$, $v_2 = 2$ which we denote as items $1, 2, 3, 4$ respectively. The corresponding lattice is given in Fig. 1. What are the nogoods for this problem? First there are those due to the explicit constraint that the two variables have distinct values: $\{v_1 = 1, v_2 = 1\}$ and $\{v_1 = 2, v_2 = 2\}$ or $\{1, 3\}$ and $\{2, 4\}$. In addition, there are necessary nogoods implied by the requirement that a variable takes on a unique value so that any set giving multiple assignments to the same variable is necessarily nogood, namely $\{v_1 = 1, v_1 = 2\}$ and $\{v_2 = 1, v_2 = 2\}$ or $\{1, 2\}$ and $\{3, 4\}$. Referring to Fig. 1, we see that these four nogoods force all sets of size 3 and 4 to be nogood too. However, sets of size zero and one are goods as are the remaining two sets of size two: $\{2, 3\}$ and $\{1, 4\}$ corresponding to $\{v_1 = 2, v_2 = 1\}$ and $\{v_1 = 1, v_2 = 2\}$ which are the solutions to this problem.

Search methods use various strategies for examining the sets in this lattice. For instance, methods such as simulated annealing (Kirkpatrick, Gelatt, & Vecchi, 1983), heuristic repair (Minton, Johnston, Philips, & Laird, 1992) and GSAT (Selman, Levesque, & Mitchell, 1992) move among complete sets, attempting to find a solution by a series of small changes to the sets. Generally these search techniques continue indefinitely if the problem has no solution and thus they can never show that a problem is insoluble. Such methods are called *incomplete*. In these methods, the search is repeated, from different initial conditions or making different random choices, until either a solution is found or some specified limit on the number of trials is reached. In the latter case, one cannot distinguish a problem with no solution at all from just a series of unlucky choices for a soluble problem. Other search techniques attempt to build solutions starting from smaller sets, often by a process of extending a consistent set until either a solution is found or no further consistent extensions are possible. In the latter case the search backtracks to a previous decision point and tries another possible extension until no further choices remain. By recording the pending choices at each decision point, these backtrack methods can determine a problem is insoluble, i.e., they are *complete* or *systematic* search methods.

---

2. The lattice of sets can also represent problems where each variable can have a different number of assigned values.

This description highlights two distinct aspects of the search procedure: a general method for moving among sets, independent of any particular problem, and a testing procedure that checks sets for consistency with the particular problem's requirements. Often, heuristics are used to make the search decisions depend on the problem structure hoping to identify changes most likely to lead to a solution and avoid unproductive regions of the search space. However, conceptually these aspects can be separated, as in the case of the quantum search algorithm presented below.

## 3. Quantum Search Methods

This section briefly describes the capabilities of quantum computers, why some straightforward attempts to exploit these capabilities for search are not particularly effective, then motivates and describes a new search algorithm.

### 3.1 An Overview of Quantum Computers

The basic distinguishing feature of a quantum computer (Benioff, 1982; Bernstein & Vazirani, 1993; Deutsch, 1985, 1989; Ekert & Jozsa, 1995; Feynman, 1986; Jozsa, 1992; Kimber, 1992; Lloyd, 1993; Shor, 1994; Svozil, 1995) is its ability to operate simultaneously on a collection of classical states, thus potentially performing many operations in the time a classical computer would do just one. Alternatively, this *quantum parallelism* can be viewed as a large parallel computer requiring no more hardware than that needed for a single processor. On the other hand, the range of allowable operations is rather limited.

To describe this more concretely, we adopt the conventional ket notation from quantum mechanics (Dirac, 1958, section 6) to denote various states[3]. That is, we use $|\alpha\rangle$ to denote the state of a computer described by $\alpha$. At a low level of description, the state of a classical computer is described by values of its bits. So for instance if it has $n$ bits, then there are $N = 2^n$ possible states for the machine, which can be associated with the numbers $s_1 = 0, \ldots, s_N = 2^n - 1$. We then say the computer is in state $|s_i\rangle$ when the values of its bits correspond to the number $i - 1$. More commonly, a computer is described in terms of higher level constructs formed from groups of bits, such as integers, character strings, sets and addresses of variables in a program. For example, a state that could arise during a search is $|\{v_1 = 1, v_2 = 1\}$, soln = False$\rangle$ corresponding to a set of assignments for variables in a CSP and a value of *false* for the program variable *soln*, e.g., used to represent whether a solution has been found. In these higher level descriptions, there will often be aspects of the computer's state, e.g., stack pointers or values for various iteration counters, that are not explicitly mentioned.

The states presented so far, where each bit or higher-level construct has a definite value, apply both to classical and quantum computers. However, quantum computers have a far richer set of possible states. Specifically, if $|s_1\rangle, \ldots, |s_N\rangle$ are the possible states for a classical computer, the possible states of the corresponding quantum computer are all linear superpositions of these states, i.e., states of the form $|s\rangle = \sum \psi_i |s_i\rangle$ where $\psi_i$ is a complex number called the *amplitude* associated with the state $|s_i\rangle$. The physical interpretation of the amplitudes comes from the measurement process. When a measurement is made on the quantum computer in state $|s\rangle$, e.g., to determine the result of the computation represented by a particular configuration of the bits in a register, one of the possible classical states is obtained. Specifically, the classical state $|s_i\rangle$ is obtained with probability $|\psi_i|^2$. Furthermore, the measurement process changes the state of the computer to exactly match the result. That is, the measurement is said to *collapse* the original superposition to the new superposition consisting of the single classical state (i.e., the amplitude of the returned state is 1 and all other amplitudes are zero). This means repeated measurements will always return the same result.

---

3. The ket notation is conceptually similar to the use of boldface to denote vectors and distinguish them from scalars.

An important consequence of this interpretation results from the fact that probabilities must sum to one. Thus the amplitudes of any superposition of states must satisfy the *normalization condition*

$$\sum_i |\psi_i|^2 = 1 \tag{2}$$

Another consequence is that the full state of a quantum computer, i.e., the superposition, is not itself an observable quantity. Nevertheless, by changing the amplitude associated with different classical states, operations on the superposition can affect the probability with which various states are observed. This possibility is crucial for exploiting quantum computation, and makes it potentially more powerful than probabilistic classical machines, in which some choices in the program are made randomly.

These superpositions can also be viewed as vectors in a space whose basis is the individual classical states $|s_i\rangle$ and $\psi_i$ is the component of the vector along the $i^{th}$ basis element of the space. Such a *state vector* can also be specified by its components as $\psi \equiv (\psi_1, \ldots, \psi_N)$ when the basis is understood from context. The inner product of two such vectors is $\phi \cdot \psi = \sum_{i=1}^{N} \phi_i^* \psi_i$ where $\phi_i^*$ denotes the complex conjugate of $\phi_i$. In matrix notation, this can also be written as $\phi^\dagger \psi$ where $\psi$ is treated as a column vector and $\phi^\dagger$ is a row vector given by the transpose of $\phi$ with all entries changed to their complex conjugate values. For these vectors, the normalization condition amounts to requiring that $\psi^\dagger \psi = 1$.

To complete this overview of quantum computers, it remains to describe how superpositions can be used within a program. In addition to the measurement process described above, there are two types of operations that can be performed on a superposition of states. The first type is to run classical programs on the machine, and the second allows for creating and manipulating the amplitudes of a superposition. In both these cases, the key property of the superposition is its linearity: an operation on a superposition of states gives the superposition of that operation acting on each of those states individually. As described below, this property, combined with the normalization condition, greatly limits the range of physically realizable operations.

In the first case, a quantum computer can perform a classical program provided it is reversible, i.e., the final state contains enough information to recover the initial state. One way to achieve this is to retain the initial input as part of the output. To illustrate the linearity of operations, consider some reversible classical computation on these states, e.g., $f(s_i)$ which produces a new state from a given input one. When applied to a superposition of states, the result is $f(|s\rangle) = \sum \psi_i |f(s_i)\rangle$. Why is reversibility required? Suppose the procedure $f$ is not reversible, i.e., it maps at least two distinct states to the same result. For example, suppose $f(s_1) = f(s_2) = s_3$. Then for the superposition $|s\rangle = \frac{1}{\sqrt{2}}(|s_1\rangle + |s_2\rangle)$ linearity requires that $f(|s\rangle) = \frac{1}{\sqrt{2}}(|f(s_1)\rangle + |f(s_2)\rangle)$ giving $\sqrt{2}|s_3\rangle$, a superposition that violates the normalization condition. Thus this irreversible classical operation is not physically realizable on a superposition, i.e., it cannot be used with quantum parallelism.

In contrast to this use of computations on individual states, the second type of operation modifies the amplitude of various states within a superposition. That is, starting from $|s\rangle = \sum \psi_k |s_k\rangle$ the operation, denoted by $U$, creates a new superposition $|s'\rangle = U|s\rangle = \sum \psi'_j |s_j\rangle$. Because the operations are linear with respect to superpositions, the new amplitudes can be expressed in terms of the original ones by $\psi'_j = \sum_k U_{jk} \psi_k$, or in matrix notation by $\psi' = U\psi$. That is, linearity means that an operation changing the amplitudes can be represented as a matrix. To satisfy the normalization condition, Eq. 2, this matrix must be such that $(\psi')^\dagger \psi' = 1$. In terms of the matrix $U$ this condition becomes[4]

$$1 = (U\psi)^\dagger(U\psi) = \psi^\dagger\left(U^\dagger U\right)\psi \tag{3}$$

which must hold for any initial state vector $\psi$ with $\psi^\dagger \psi = 1$. To see what this implies about the matrix $A \equiv U^\dagger U$, suppose $\psi = \hat{\mathbf{e}}_j = (\dots, 0, 1, 0, \dots)$ is the $j^{th}$ unit vector, corresponding to the superposition $|s_j\rangle$ where all amplitudes are zero except for $\psi_j = 1$. In this case $\psi^\dagger A \psi = A_{jj}$ which must equal one by Eq. 3. That is, the diagonal elements of $U^\dagger U$ must all be equal to one. For $\psi = \frac{1}{\sqrt{2}}(\hat{\mathbf{e}}_j + \hat{\mathbf{e}}_k)$ with $j \neq k$,

$$\begin{aligned}
\psi^\dagger A \psi &= \frac{1}{2}(\hat{\mathbf{e}}_j + \hat{\mathbf{e}}_k)A(\hat{\mathbf{e}}_j + \hat{\mathbf{e}}_k) \\
&= \frac{1}{2}[A_{jj} + A_{kk} + A_{jk} + A_{kj}]
\end{aligned} \tag{4}$$

This must equal one by Eq. 3, and we already know that the diagonal terms equal one. Thus we conclude $A_{jk} = -A_{kj}$. A similar argument using $\psi = \frac{1}{\sqrt{2}}(\hat{\mathbf{e}}_j + i\hat{\mathbf{e}}_k)$, a superposition with an imaginary value for the second amplitude, gives $A_{jk} = A_{kj}$. Together these conditions mean that $A$ is the identity matrix, so $U^\dagger U = I$, i.e., the matrix $U$ must be unitary to operate on superpositions. Moreover, this condition is sufficient to make *any* initial state satisfy Eq. 3. This shows how the restriction to linear unitary operations arises directly from the linearity of quantum mechanics and Eq. 2, the normalization condition for probabilities. The class of unitary matrices includes permutations, rotations and arbitrary phase changes (i.e., diagonal matrices where each element on the diagonal is a complex number with magnitude equal to one).

Reversible classical programs, unitary operations on the superpositions and the measurement process are the basic ingredients used to construct a program for a quantum computer. As used in the search algorithm described below, such a program consists of first preparing an initial superposition of states, operating on those states with a series of unitary matrices in conjunction with a classical program to evaluate the consistency of various states with respect to the search requirements, and then making a measurement to obtain a definite final answer. The amplitudes of the superposition just before the measurement is made determine the probability of obtaining a solution. The overall structure is a probabilistic Monte Carlo computation (Motwani & Raghavan, 1995) in which at each trial there is some probability to get a solution, but no guarantee. This means the search method is incomplete: it can find a solution if one exists but can never guarantee a solution doesn't exist.

---

4. $U^\dagger$ is the transpose of $U$ with all elements changed to their complex conjugates. That is $\left(U^\dagger\right)_{jk} = (U_{kj})^\star$.

An alternate conceptual view of these quantum programs is provided by the path integral approach to quantum mechanics (Feynman, 1985). In this view, the final amplitude of a given state is obtained by a weighted sum over all possible paths that produce that state. In this way, the various possibilities involved in a computation can interfere with each other, either constructively or destructively. This differs from the classical combination of probabilities of different ways to reach the same outcome (e.g., as used in probabilistic algorithms): the probabilities are simply added, giving no possibility for interference. Interference is also seen in classical waves, such as with sound or ripples on the surface of water. But these systems lack the capability of quantum parallelism. The various formulations of quantum mechanics, involving operators, matrices or sums over paths are equivalent but suggest different intuitions about constructing possible quantum algorithms.

## 3.2 Example: A One-Bit Computer

A simple example of these ideas is given by a single bit. In this case there are two possible classical states $|0\rangle$ and $|1\rangle$ corresponding to the values 0 and 1, respectively, for the bit. This defines a two dimensional vector space of superpositions for a quantum bit. There are a number of proposals for implementing quantum bits, i.e., devices whose quantum mechanical properties can be controlled to produce desired superpositions of two classical values. One example (DiVincenzo, 1995; Lloyd, 1995) is an atom whose ground state corresponds to the value 0 and an excited state to the value 1. The use of lasers of appropriate frequencies can switch such an atom between the two states or create superpositions of the two classical states. This ability to manipulate quantum superpositions has been demonstrated in small cases (Zhu, Kleiman, Li, Lu, Trentelman, & Gordon, 1995). Another possibility is through the use of atomically precise manipulations (DiVincenzo, 1995) using a scanning tunneling or atomic force microscope. This possibility of precise manipulation of chemical reactions has also been demonstrated (Muller, Klein, Lee, Clarke, McEuen, & Schultz, 1995). There are also a number of other proposals under investigation (Barenco, Deutsch, & Ekert, 1995; Sleator & Weinfurter, 1995; Cirac & Zoller, 1995), including the possibility of multiple simultaneous quantum operations (Margolus, 1990).

A simple computation on a quantum bit is the logical NOT operation, i.e., $\text{NOT}(|0\rangle) = |1\rangle$ and $\text{NOT}(|1\rangle) = |0\rangle$. This operator simply exchanges the state vector's components:

$$\text{NOT}\begin{pmatrix} \psi_0 \\ \psi_1 \end{pmatrix} \equiv \text{NOT}(\psi_0|0\rangle + \psi_1|1\rangle) = \psi_0|1\rangle + \psi_1|0\rangle \equiv \begin{pmatrix} \psi_1 \\ \psi_0 \end{pmatrix} \tag{5}$$

This operation can also be represented as multiplication by the permutation matrix $\begin{pmatrix} 0 & 1 \\ 1 & 0 \end{pmatrix}$. Another operator is given by the rotation matrix

$$U(\theta) = \begin{pmatrix} \cos\theta & -\sin\theta \\ \sin\theta & \cos\theta \end{pmatrix} \tag{6}$$

This can be used to create superpositions from single classical states, e.g.,

$$U\left(\frac{\pi}{4}\right)\begin{pmatrix} 1 \\ 0 \end{pmatrix} \equiv U\left(\frac{\pi}{4}\right)|0\rangle = \frac{1}{\sqrt{2}}(|0\rangle + |1\rangle) \equiv \frac{1}{\sqrt{2}}\begin{pmatrix} 1 \\ 1 \end{pmatrix} \tag{7}$$

This rotation matrix can also be used to illustrate interference, an important way in which quantum computers differ from probabilistic classical algorithms. First, consider a classical algorithm with two methods for generating random bits, $R_0$ (producing a "0" with probability 3/4) and $R_1$ (producing a "0" with probability 1/4). Suppose a "0" represents a failure (e.g., a probabilistic search that does not find a solution) while "1" represents a success. Finally, let the classical algorithm consist of selecting one of these methods to use, with probability $p$ to pick $R_0$. Then the overall probability to obtain a "0" as the final result is just $\frac{3}{4}p + \frac{1}{4}(1-p)$ or

$$P_{classical} = \frac{1}{4} + \frac{p}{2} \tag{8}$$

The best that can be done is to choose $p = 0$, giving a probability of 1/4 for failure.

A quantum analog of this simple calculation can be obtained from a rotation with $\theta = \frac{\pi}{3}$. Starting from the individual classical states this gives superpositions

$$\begin{aligned} U\left(\frac{\pi}{3}\right)\begin{pmatrix} 1 \\ 0 \end{pmatrix} &= \frac{1}{2}\begin{pmatrix} \sqrt{3} \\ 1 \end{pmatrix} \\ U\left(\frac{\pi}{3}\right)\begin{pmatrix} 0 \\ 1 \end{pmatrix} &= \frac{1}{2}\begin{pmatrix} -1 \\ \sqrt{3} \end{pmatrix} \end{aligned} \tag{9}$$

which correspond to the generators $R_0$ and $R_1$ respectively, because of their respective probabilities of 3/4 and 1/4 to produce a "0" when measured. Starting instead from a superposition of the two classical states, $\begin{pmatrix} \cos\phi \\ \sin\phi \end{pmatrix}$, corresponds to the step of the classical algorithm where generator $R_0$ is selected with probability $p = \cos^2\phi$. The resulting state after applying the rotation, $U(\frac{\pi}{3})\begin{pmatrix} \cos\phi \\ \sin\phi \end{pmatrix}$, has probability

$$\begin{aligned} P_{quantum} &= \frac{1}{4} + \frac{\cos^2\phi}{2} - \frac{\sqrt{3}}{4}\sin(2\phi) \\ &= P_{classical} - \frac{\sqrt{3}}{4}\sin(2\phi) \end{aligned} \tag{10}$$

to produce a "0" value. In this case the minimum value of the probability to obtain a "0" is not 1/4 but in fact can be made to equal 0 with the choice $\phi = \frac{\pi}{3}$. In this case the amplitudes from the two original states exactly cancel each other, an example of destructive interference.

As a final example, illustrating the limits of operations on superpositions, consider the simple classical program that sets a bit to the value one. That is, SET($|0\rangle$) = $|1\rangle$ and $\text{SET}(|1\rangle) = |1\rangle$. This operation is not reversible: knowing the result does not determine the original input. By linearity, $\text{SET}\left(\frac{1}{\sqrt{2}}(|0\rangle + |1\rangle)\right) = \frac{1}{\sqrt{2}}(\text{SET}(|0\rangle) + \text{SET}(|1\rangle))$, which in turn is $\frac{1}{\sqrt{2}}2|1\rangle = \sqrt{2}|1\rangle$. This state violates the normalization condition. Thus we see that this classical operation is not physically realizable for a quantum computer. Similarly, another common classical operation, making a copy of a bit, is also ruled out (Svozil, 1995), forming the basis for quantum cryptography (Bennett, 1992).

### 3.3 Some Approaches to Search

A device consisting of $n$ quantum bits allows for operations on superpositions of $2^n$ classical states. This ability to operate simultaneously on an exponentially large number of states with just a linear number of bits is the basis for quantum parallelism. In particular, repeating the operation of Eq. 7 $n$ times, each on a different bit, gives a superposition with equal amplitudes in $2^n$ states.

At first sight quantum computers would seem to be ideal for combinatorial search problems that are in the class NP. In such problems, there is an efficient procedure $f(s)$ that takes a potential solution set $s$ and determines whether $s$ is in fact a solution, but there are exponentially many potential solutions, very few of which are in fact solutions. If $s_1, \ldots, s_N$ are the potential sets to consider, we can quickly form the superposition $\frac{1}{\sqrt{N}}(|s_1\rangle + \ldots + |s_N\rangle)$ and then simultaneously evaluate $f(s)$ for all these states, resulting in a superposition of the sets and their evaluation, i.e., $\frac{1}{\sqrt{N}}\sum |s_i, \text{soln} = f(s_i)\rangle$. Here $|s_i, \text{soln} = f(s_i)\rangle$ represents a classical search state considering the set $s_i$ along with a variable *soln* whose value is true or false according to the result of evaluating the consistency of the set with respect to the problem requirements. At this point the quantum computer has, in a sense, evaluated all possible sets and determined which are solutions. Unfortunately, if we make a measurement of the system, we get each set with equal probability $1/N$ and so are very unlikely to observe a solution. This is thus no better than the slow classical search method of random generate-and-test where sets are randomly constructed and tested until a solution is found. Alternatively, we can obtain a solution with high probability by repeating this operation $O(N)$ times, either serially (taking a long time) or with multiple copies of the device (requiring a large amount of hardware or energy if, say, the computation is done by using multiple photons). This shows a trade-off between time and energy (or other physical resources), conjectured to apply more generally to solving these search problems (Cerny, 1993), and also seen in the trade-off of time and number of processors in parallel computers.

To be useful for combinatorial search, we can't just evaluate the various sets but instead must arrange for amplitude to be concentrated into the solution sets so as to greatly increase the probability a solution will be observed. Ideally this would be done with a mapping that gives constructive interference of amplitude in solutions and destructive interference in non-solutions. Designing such maps is complicated by the fact that they must be linear unitary operators as described above. Beyond this physical restriction, there is an algorithmic or computational requirement: the mapping should be efficiently computable (DiVincenzo & Smolin, 1994). For example, the map cannot require a priori knowledge of the solutions (otherwise constructing the map would require first doing the search). This computational requirement is analogous to the restriction on search heuristics: to be useful, the heuristic itself must not take a long time to compute. These requirements on the mapping trade off against each other. Ideally one would like to find a way to satisfy them all so the map can be computed in polynomial time and give, at worst, polynomially small probability to get a solution if the problem is soluble. One approach is to arrange for constructive interference in solutions while nonsolutions receive random contributions to their amplitude. While such random contributions are not as effective as a complete destructive interference, they are easier to construct and form the basis for a recent factoring algorithm (Shor, 1994) as well as the method presented here.

Classical search algorithms can suggest ways to combine the use of superpositions with interference. These include local repair styles of search where complete assignments are modified, and backtracking search, where solutions are built up incrementally. Using superpositions, many possibilities could be simultaneously considered. However these search methods have no a priori specification of the number of steps required to reach a solution so it is unclear how to determine when enough amplitude might be concentrated into solution states to make a measurement worthwhile. Since the measurement process destroys the superposition, it is not possible to resume the computation at the point where the measurement was made if it does not produce a solution. A more subtle problem arises because different search choices lead to solutions in differing numbers of steps. Thus one would also need to maintain any amplitude already in solution states while the search continues. This is difficult due to the requirement for reversible computations.

While it may be fruitful to investigate these approaches further, the quantum method proposed below is based instead on a breadth-first search that incrementally builds up all solutions. Classically, such methods maintain a list of goods of a given size. At each step, the list is updated to include all goods with one additional variable. Thus at step $i$, the list consists of sets of size $i$ which are used to create the new list of sets of size $i + 1$. For a CSP with $n$ variables, $i$ ranges from 0 to $n - 1$, and after completing these $n$ steps the list will contain all solutions to the problem. Classically, this is not a useful method for finding a single solution because the list of partial assignments grows exponentially with the number of steps taken. A quantum computer, on the other hand, can handle such lists readily as superpositions. In the method described below, the superposition at step $i$ consists of all sets of size $i$, not just consistent ones, i.e., the sets at level $i$ in the lattice. There is no question of when to make the final measurement because the computation requires exactly $n$ steps. Moreover, there is an opportunity to use interference to concentrate amplitude toward goods. This is done by changing the phase of amplitudes corresponding to nogoods encountered while moving through the lattice.

As with the division of search methods into a general strategy (e.g., backtrack) and problem specific choices, the quantum mapping described below has a general matrix that corresponds to exploring all possible changes to the partial sets, and a separate, particularly simple, matrix that incorporates information on the problem specific constraints. More complex maps are certainly possible, but this simple decomposition is easier to design and describe. With this decomposition, the difficult part of the quantum mapping is independent of the details of the constraints in a particular problem. This suggests the possibility of implementing a special purpose quantum device to perform the general mapping. The constraints of a specific problem are used only to adjust phases as described below, a comparatively simple operation.

For constraint satisfaction problems, a simple alternative representation to the full lattice structure is to use partial assignments only, i.e., sets of variable-value pairs that have no variable more than once. At first sight this might seem better in that it removes from consideration the necessary nogoods and hence increases the proportion of complete sets that are solutions. However, in this case the number of sets as a function of level in the lattice decreases before reaching the solution level, precluding the simple form of a unitary mapping described below for the quantum search algorithm. Another representation that avoids this problem is to consider assignments in only a single order for the variables (selected randomly or through the use of heuristics). This version of the set lattice has been previously used in theoretical analyses of phase transitions in search (Williams & Hogg, 1994). This may be useful to explore further for the quantum search, but is unlikely to be as effective. This is because in a fixed ordering some sets will become nogood only at the last few steps, resulting is less opportunity for interference based on nogoods to focus on solutions.

## 3.4 Motivation

To motivate the mapping described below, we consider an idealized version. It shows why paths through the lattice tend to interfere destructively for nonsolution states, provided the constraints are small.

The idealized map simply maps each set in the lattice equally to its supersets at the next level, while introducing random phases for sets found to be nogood. For this discussion we are concerned with the relative amplitude in solutions and nogoods so we ignore the overall normalization. Thus for instance, with $N = 6$, the state $|\{1, 2\}\rangle$ will map to an unnormalized superposition of its four supersets of size 3, namely the state $|\{1, 2, 3\}\rangle + \ldots + |\{1, 2, 6\}\rangle$.

With this mapping, a good at level $j$ will receive equal contribution from each of its $j$ subsets at the prior level. Starting with amplitude of 1 at level 0 then gives an amplitude of $j!$ for goods at level $j$. In particular, $L!$ for solutions.

How does this compare with contribution to nogoods, on average? This will depend on how many of the subsets are nogoods also. A simple case for comparison is when *all* sets in the lattice are nogood (starting with those at level $k$ given by the size of the constraints, e.g., $k = 2$ for problems with binary constraints). Let $r_j$ be the expected value of the magnitude of the amplitude for sets at level $j$. Each set at level $k$ will have $r_k = k!$ (and a zero phase) because all smaller subsets will be goods. A set $s$ at level $j > k$ will be a sum of $j$ contributions from (nogood) subsets, giving a total contribution of

$$\psi(s) = \sum_{m=1}^{j} \psi(s_m) e^{i\phi_m} \tag{11}$$

where the $s_m$ are the subsets of $s$ of size $j - 1$ and the phases $\phi_m$ are randomly selected. The $\psi(s_m)$ have expected magnitude $r_{j-1}$ and some phase that can be combined with $\phi_m$ to give a new random phase $\theta_m$. Ignoring the variation in the magnitude of the amplitudes at each level this gives

$$r_j = r_{j-1} \left\langle \sum_{m=1}^{j} e^{i\theta_m} \right\rangle = r_{j-1} \sqrt{j} \tag{12}$$

because the sum of $j$ random phases is equivalent to an unbiased random walk (Karlin & Taylor, 1975) with $j$ unit steps which has expected net distance of $\sqrt{j}$. Thus $r_j = r_k\sqrt{j!/k!}$ or $r_j = \sqrt{j!k!}$ for $j > k$.

This crude argument gives a rough estimate of the relative probabilities for solutions compared to complete nogoods. Suppose there is only one solution. Then its relative probability is $L!^2$. The nogoods have relative probability $(N_L - 1)r_L^2 \sim N_L L! k!$ with $N_L$ given by Eq. 1. An interesting scaling regime is $L = N/b$ with fixed $b$, corresponding to a variety of well-studied constraint satisfaction problems. This gives

$$\ln\left(\frac{P_{soln}}{P_{nogood}}\right) = \ln\left(\frac{L!}{N_L k!}\right) \sim \frac{N}{b}\ln N + O(N) \tag{13}$$

This goes to infinity as problems get large so the enhancement of solutions is more than enough to compensate for their rareness among sets at the solution level.

The main limitation of this argument is assuming that all subsets of a nogood are also nogood. For many nogoods, this will not be the case, resulting in less opportunity for cancellation of phases. The worst situation in this respect is when most subsets are goods. This could be because the constraints are large, i.e., they don't rule out states until many items are included. Even with small constraints, this could happen occasionally due to a poor ordering choice for adding items to the sets, hence suggesting that a lattice restricted to assignments in a single order will be much less effective in canceling amplitude in nogoods. For the problems considered here, with small constraints, a large nogood cannot have too many good subsets because to be nogood means a small subset violates a (small) constraint and hence most subsets obtained by removing one element will still contain that bad subset giving a nogood. In fact, some numerical experiments (with the class of unstructured problems described below) show that this mapping is very effective in canceling amplitude in the nogoods. Thus the assumptions made in this simplified argument seem to provide the correct intuitive description of the behavior.

Still the assumption of many nogood subsets underlying the above argument does suggest the extreme cancellation derived above will *least* apply when the problem has many large partial solutions. This gives a simple explanation for the difficulty encountered with the full map described below at the phase transition point: this transition is associated with problems with relatively many large partial solutions but few complete solutions. Hence we can expect relatively less cancellation of at least some nogoods at the solution level and a lower overall probability to find a solution.

This discussion suggests why a mapping of sets to supersets along with random phases introduced at each inconsistent set can greatly decrease the contribution to nogoods at the solution level. However, this mapping itself is not physically realizable because it is not unitary. For example, the mapping from level 1 to 2 with $N = 3$ takes the states $|\{1\}\rangle, |\{2\}\rangle, |\{3\}\rangle$ to $|\{1,2\}\rangle, |\{1,3\}\rangle, |\{2,3\}\rangle$ with the matrix

$$M = \begin{pmatrix} 1 & 1 & 0 \\ 1 & 0 & 1 \\ 0 & 1 & 1 \end{pmatrix} \tag{14}$$

Here, the first column means the state $|\{1\}\rangle$ contributes equally to $|\{1,2\}\rangle$ and $|\{1,3\}\rangle$, its supersets, and gives no contribution to $|\{2,3\}\rangle$. We see immediately that the columns of this matrix are not orthogonal, though they can be easily normalized by dividing the entries by $\sqrt{2}$. More generally, this mapping takes each set at level $i$ to the $N - i$ sets with one more element. The corresponding matrix $M$ has one column for each $i$-set and one row for each $(i + 1)$-set. In each column there will be exactly $N - i$ 1's (corresponding to the supersets of the given $i$-set) and the remaining entries will be 0. Two columns will have at most a single nonzero value in common (and only when the two corresponding $i$-sets have all but one of their values in common: this is the only way they can share a superset in common). This means that as $N$ gets large, the columns of this matrix are almost orthogonal (provided $i < N/2$, the case of interest here). This fact is used below to obtain a unitary matrix that is fairly close to $M$.

## 3.5   A Search Algorithm

The general idea of the mapping introduced here is to move as much amplitude as possible to supersets (just as in classical breadth-first search, increments to partial sets give supersets). This is combined with a problem specific adjustment of phases based on testing partial states for consistency (this corresponds to a diagonal matrix and thus is particularly simple in that it does not require any mixing of the amplitudes of different states). The specific methods used are described in this section.

### 3.5.1   The Problem-Independent Mapping

To take advantage of the potential cancellation of amplitude in nogoods described above we need a unitary mapping whose behavior is similar to the ideal mapping to supersets. There are two general ways to adjust the ideal mapping of sets to supersets (mixtures of these two approaches are possible as well). First, we can keep some amplitude at the same level of the lattice instead of moving all the amplitude up to the next level. This allows using the ideal map described above (with suitable normalization) and so gives excellent discrimination between solutions and nonsolutions, but unfortunately not much amplitude reaches solution level. This is not surprising: the use of random phases cancel the amplitude in nogoods but this doesn't add anything to solutions (because solutions are not a superset of any nogood and hence cannot receive any amplitude from them). Hence at best, even when all nogoods cancel completely, the amplitude in solutions will be no more than their relative number among complete sets, i.e., very small. Thus the random phases prevent much amplitude moving to nogoods high in the lattice, but instead of contributing to solutions this amplitude simply remains at lower levels of the lattice. Hence we have no better chance than random selection of finding a solution (but, when a solution is not found, instead of getting a nogood at the solution level, we are now likely to get a smaller set in the lattice). Thus we must arrange for amplitude taken from nogoods to contribute instead to the goods. This requires the map to take amplitude to sets other than just supersets, at least to some extent.

The second way to fix the nonunitary ideal map is to move amplitude also to non-supersets. This can move all amplitude to the solution level. It allows some canceled amplitude from nogoods to go to goods, but also vice versa, resulting in less effective concentration into solutions. This can be done with a unitary matrix as close as possible to the nonunitary ideal map to supersets, and that also has a relatively simple form. The general question here is given $k$ linearly independent vectors in $m$ dimensional space, with $k \leq m$, find $k$ orthonormal vectors in the space as close as possible to the $k$ original ones. Restricting attention to the subspace defined by the original vectors, this can be obtained[5] using the singular value decomposition (Golub & Loan, 1983) (SVD) of the matrix $M$ whose columns are the $k$ given vectors. Specifically, this decomposition is $M = A^\dagger \Lambda B$, where $\Lambda$ is a diagonal matrix containing the singular values of $M$ and both $A^\dagger$ and $B$ have orthonormal columns. For a real matrix $M$, the matrices of the decomposition are also real-valued. The matrix $U = A^\dagger B$ has orthonormal columns and is the closest set of orthogonal vectors according to the Frobenius matrix norm. That is, this choice for $U$ minimizes $|U - M|^2 \equiv \sum_{rs} |U_{rs} - M_{rs}|^2$ among all unitary matrices. This construction fails if $k > m$ since an $m$–dimensional space cannot have more than $m$ orthogonal vectors. Hence we restrict consideration to mappings in the lattice at those levels $i$ where level $i+1$ has at least as many sets as level $i$, i.e., $N_i \leq N_{i+1}$. Obtaining a solution requires mapping up to level $L$ so, from Eq. 1, this restricts consideration to problems where $L \leq \lceil N/2 \rceil$.

For example, the mapping from level 1 to 2 with $N = 3$ given in Eq. 14 has the singular value decomposition $M = A^\dagger \Lambda B$ with this decomposition given explicitly as

$$A^\dagger \Lambda B = \begin{pmatrix} \frac{1}{\sqrt{3}} & -\frac{1}{\sqrt{2}} & \frac{1}{\sqrt{6}} \\ \frac{1}{\sqrt{3}} & \frac{1}{\sqrt{2}} & \frac{1}{\sqrt{6}} \\ \frac{1}{\sqrt{3}} & 0 & -\sqrt{\frac{2}{3}} \end{pmatrix} \begin{pmatrix} 2 & 0 & 0 \\ 0 & 1 & 0 \\ 0 & 0 & 1 \end{pmatrix} \begin{pmatrix} \frac{1}{\sqrt{3}} & \frac{1}{\sqrt{3}} & \frac{1}{\sqrt{3}} \\ 0 & -\frac{1}{\sqrt{2}} & \frac{1}{\sqrt{2}} \\ \sqrt{\frac{2}{3}} & -\frac{1}{\sqrt{6}} & -\frac{1}{\sqrt{6}} \end{pmatrix} \tag{15}$$

The closest unitary matrix is then

$$U = A^\dagger B = \frac{1}{3} \begin{pmatrix} 2 & 2 & -1 \\ 2 & -1 & 2 \\ -1 & 2 & 2 \end{pmatrix} \tag{16}$$

While this gives a set of orthonormal vectors close to the original map, one might be concerned about the requirement to compute the SVD of exponentially large matrices. Fortunately, however, the resulting matrices have a particularly simple structure in that the entries depend only on the overlap between the sets. Thus we can write the matrix elements in the form $U_{r\beta} = a_{|r \cap \beta|}$ ($r$ is an (i+1)-subset, $\beta$ is an i-subset). The overlap $|r \cap \beta|$ ranges from $i$ when $\beta \subset r$ to 0 when there is no overlap. Thus instead of exponentially many distinct values, there are only $i + 1$, a linear number. This can be exploited to give a simpler method for evaluating the entries of the matrix as follows.

We can get expressions for the $a$ values for a given $N$ and $i$ since the resulting column vectors are orthonormal. Restricting attention to real values, this gives

$$1 = \left( U^\dagger U \right)_{\alpha\alpha} = \sum_{k=0}^{i} n_k a_k^2 \tag{17}$$

where

$$n_k = \binom{i}{k} \binom{N - i}{i + 1 - k} \tag{18}$$

---

5. I thank J. Gilbert for pointing out this technique, as a variant of the orthogonal Procrustes problem (Golub & Loan, 1983).

is the number of ways to pick $r$ with the specified overlap. For the off-diagonal terms, suppose $|\alpha \cap \beta| = p < i$ then, for real values of the matrix elements,

$$0 = \left( U^\dagger U \right)_{\alpha\beta} = \sum_{j,k=0}^{i} n_{jk}^{(p)} a_j a_k \tag{19}$$

where

$$n_{jk}^{(p)} = \sum_x \binom{i-p}{k-x}\binom{p}{x}\binom{i-p}{j-x}\binom{N-2i+p}{i+1-j-k+x} \tag{20}$$

is the number of sets $r$ with the required overlaps with $\alpha$ and $\beta$, i.e., $|r \cap \alpha| = k \leq i$ and $|r \cap \beta| = j \leq i$. In this sum, $x$ is the number of items the set $r$ has in common with both $\alpha$ and $\beta$. Together these give $i + 1$ equations for the values of $a_0, \ldots, a_i$, which are readily solved numerically[6]. There are multiple solutions for this system of quadratic equations, each representing a possible unitary mapping. But there is a unique one closest to the ideal mapping to supersets, as given by the SVD. It is this solution we use for the quantum search algorithm[7], although it is possible some other solution, in conjunction with various choices of phases, performs better. Note that the number of values and equations grows only linearly with the level in the lattice, even though the number of sets at each level grows exponentially. When necessary to distinguish the values at different levels in the lattice, we use $a_k^{(i)}$ to mean the value of $a_k$ for the mapping from level $i$ to $i + 1$.

The example of Eq. 14, with $N = 3$ and $i = 1$, has $1 = a_0^2 + 2a_1^2$ for Eq. 17 and $0 = 2a_0 a_1 + a_1^2$ for Eq. 19. The solution of these unitarity conditions closest to Eq. 14 is $a_0 = -\frac{1}{3}$, $a_1 = \frac{2}{3}$ corresponding to Eq. 16.

A normalized version of the ideal map has $a_i^{(i)} = \frac{1}{\sqrt{n_i}} = \frac{1}{\sqrt{N-i}}$ and all other values equal to zero. The actual values for $a_k^{(i)}$ are fairly close to this (confirming that the ideal map is close to orthogonal already), and alternate in sign. To illustrate their behavior, it is useful to consider the scaled values $b_k^{(i)} \equiv (-1)^k a_{i-k}^{(i)} \sqrt{n_{i-k}}$, with $n_{i-k}$ evaluated using Eq. 18. The behavior of these values for $N = 10$ is shown in Fig. 2. Note that $b_0^{(i)}$ is close to one, and decreases slightly as higher levels in the lattice (i.e., larger $i$ values) are considered: the ideal mapping is closer to orthogonal at low levels in the lattice.

Despite the simple values for the example of Eq. 16, the $a_k$ values in general do not appear to have a simple closed form expression. This is suggested by obtaining exact solutions to Eqs. 17 and 19 using the Mathematica symbolic algebra program (Wolfram, 1991). In most cases this gives complicated expressions involving nested roots. Since such expressions could simplify, the $a_k$ values were also checked for being close to rational numbers and whether they are roots of single variable polynomials of low degree[8]. Neither simplification was found to apply.

Finally we should note that this mapping only describes how the sets at level $i$ are mapped to the next level. The full quantum system will also perform some mapping on the remaining sets in the lattice. By changing the map at each step, most of the other sets can simply be left unchanged, but there will need to be a map of the sets at level $i + 1$ other than the identity mapping to be orthogonal to the map from level $i$. Any orthogonal set mapping partly back to level $i$ and partly remaining in sets at level $i + 1$ will be suitable for this: in our application there is no amplitude at level $i + 1$ when the map is used and hence it doesn't matter what mapping is used. However, the choice of this part of the overall mapping remains a degree of freedom that could perhaps be exploited to minimize errors introduced by external noise.

---

6. High precision values were obtained from the FindRoot function of Mathematica.

7. The values are given in Online Appendix 1.

8. Using the Mathematica function Rationalize and the package NumberTheory`Recognize`.

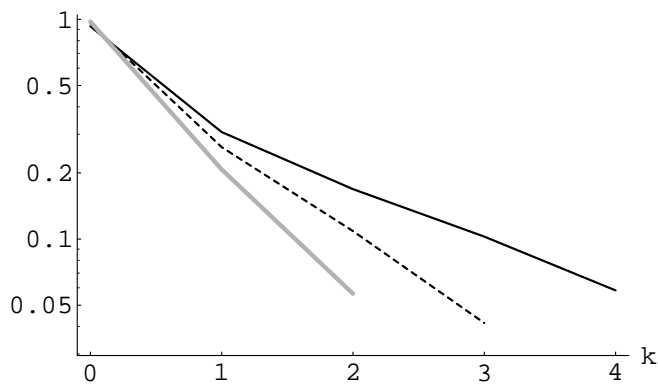

Figure 2: Behavior of $b_k^{(i)}$ vs. $k$ on a log scale for $N = 10$. The three curves show the values for $i = 4$ (black), 3 (dashed) and 2 (gray).

### 3.5.2 Phases for Nogoods

In addition to the general mapping from one level to the next, there is the problem-specific aspect of the algorithm, namely the choice of phases for the nogood sets at each level. In the ideal case described above, random phases were given to each nogood, resulting in a great deal of cancellation for nogoods at the solution level. While this is a reasonable choice for the unitary mapping, other policies are possible as well. For example, one could simply invert the phase of each nogood[9] (i.e., multiply its amplitude by -1). This choice doesn't work well for the idealized map to supersets only but, as shown below, is helpful for the unitary map. It can be motivated by considering the coefficients shown in Fig. 2. Specifically, when a nogood is encountered for the first time on a path through the lattice, we would like to cancel phase to its supersets but at the same time enhance amplitude in other sets likely to lead to solutions. Because $a_{i-1}^{(i)}$ is negative, inverting the phase will tend to add to sets that differ by one element from the nogood. At least some of these will avoid violating the small constraint that produced this nogood set, and hence may contribute eventually to sets that do lead to solutions.

Moreover, one could use information on the sets at the next level to decide what to do with the phase: as currently described, the computation makes no use of testing the consistency of sets at the solution level itself, and hence is completely ineffective for problems where the test requires the complete set. Perhaps better would be to mark a state as nogood if it has no consistent extensions with one more item (this is simple to check since the number of extensions grows only linearly with problem size). Another possibility is for the phase to be adjusted based on how many constraints are violated, which could be particularly appropriate for partial constraint satisfaction problems (Freuder & Wallace, 1992) or other optimization searches.

---

9. I thank J. Lamping for suggesting this.

### 3.5.3 SUMMARY

The search algorithm starts by evenly dividing amplitude among the goods at a low level $K$ of the lattice. A convenient choice for binary CSPs is to start at level $K = 2$, where the number of sets is proportional to $N^2$. Then for each level from $K$ to $L - 1$, we adjust the phases of the states depending on whether they are good or nogood and map to the next level. Thus if $\psi_\alpha^{(j)}$ represents the amplitude of the set $\alpha$ at level $j$, we have

$$\psi_r^{(j+1)} = \sum_\alpha U_{r\alpha} \rho_\alpha \psi_\alpha^{(j)} = \sum_k a_k^{(j)} \sum_{|r \cap \alpha| = k} \rho_\alpha \psi_\alpha^{(j)} \tag{21}$$

where $\rho_\alpha$ is the phase assigned to the set $\alpha$ after testing whether it is nogood, and the final inner sum is over all sets $\alpha$ that have $k$ items in common with $r$. That is, $\rho_\alpha = 1$ when $\alpha$ is a good set. For nogoods, $\rho_\alpha = -1$ when using the phase inversion method, and $\rho_\alpha = e^{i\theta}$ with $\theta$ uniformly selected from $[0, 2\pi)$ when using the random phase method. Finally we measure the state, obtaining a complete set. This set will be a solution with probability

$$p_{soln} = \sum_s \left| \psi_s^{(L)} \right|^2 \tag{22}$$

with the sum over solution sets, depending on the particular problem and method for selecting the phases.

What computational resources are required for this algorithm? The storage requirements are quite modest: $N$ bits can produce a superposition of $2^N$ states, enough to represent all the possible sets in the lattice structure. Since each trial of this algorithm gives a solution only with probability $p_{soln}$, on average it will need to be repeated $1/p_{soln}$ times to find a solution. The cost of each trial consists of the time required to construct the initial superposition and then evaluate the mapping on each step from the level $K$ to the solution level $L$, a total of $L - K < N/2$ mappings. Because the initial state consists of sets of size $K$, there are only a polynomial number of them (i.e., $O\left(N^K\right)$) and hence the cost to construct the initial superposition will be relatively modest. The mapping from one level to the next will need to be produced by a series of more elementary operations that can be directly implemented in physical devices. Determining the required number of such operations remains an open question, though the particularly simple structure of the matrices should not require involved computations and should also be able to exploit special purpose hardware. At any rate, this mapping is independent of the structure of the problem and its cost does not affect the relative costs of different problem structures. Finally, determining the phases to use for the nogood sets involves testing the sets against the constraints, a relatively rapid operation for NP search problems. Thus to examine how the cost of this search algorithm depends on problem structure, the key quantity is the behavior of $p_{soln}$.

### 3.6   An Example of Quantum Search

To illustrate the algorithm's operation and behavior, consider the small case of $N = 3$ with the map starting from level $K = 0$ and going up to level $L = 2$. Suppose that $\{3\}$ and its supersets are the only nogoods. We begin with all amplitude in the empty set, i.e., with the state $|\emptyset\rangle$. The map from level 0 to 1 gives equal amplitude to all singleton sets, producing $\frac{1}{\sqrt{3}}(|\{1\}\rangle + |\{2\}\rangle + |\{3\}\rangle)$. We then introduce a phase for the nogood set, giving $\frac{1}{\sqrt{3}}\Big(|\{1\}\rangle + |\{2\}\rangle + e^{i\theta}|\{3\}\rangle\Big)$. Finally we use Eq. 16 to map this to the sets at level 2, giving the final state

$$\frac{1}{3\sqrt{3}}\Big(\big(4 - e^{i\theta}\big)|\{1,2\}\rangle + \big(1 + 2e^{i\theta}\big)|\{1,3\}\rangle + \big(1 + 2e^{i\theta}\big)|\{2,3\}\rangle\Big) \tag{23}$$

At this level, only set $\{1,2\}$ is good, i.e., a solution. Note that the algorithm does not make any use of testing the states at the solution level for consistency.

The probability to obtain a solution when the final measurement is made is determined by the amplitude of the solution level, so in this case Eq. 22 becomes

$$p_{soln} = \left|\frac{1}{3\sqrt{3}}\Big(4 - e^{i\theta}\Big)\right|^2 = \frac{1}{27}(17 - 8\cos\theta) \tag{24}$$

From this we can see the effect of different methods for selecting the phase for nogoods. If the phase is selected randomly, $p_{soln} = \frac{17}{27} = 0.63$ because the average value of $\cos\theta$ is zero. Inverting the phase of the nogood, i.e., using $\theta = \pi$, gives $p_{soln} = \frac{25}{27} = 0.93$. These probabilities compare with the $1/3$ chance of selecting a solution by random choice. In this case, the optimal choice of phase is the same as that obtained by simple inversion. However this is not true in general: picking phases optimally will require knowledge about the solutions and hence is not a feasible mapping. Note also that even the optimal choice of phase doesn't guarantee a solution is found.

## 4.   Average Behavior of the Algorithm

In this section, the behavior of the quantum algorithm is evaluated for two classes of combinatorial search problems. The first class, of unstructured problems, is used to examine the phase transition in a particularly simple context using both random and inverted phases for nogoods. The second class, random propositional satisfiability (SAT), evaluates the robustness of the algorithm for problems with particular structure.

For classical simulation of this algorithm we explicitly compute the amplitude of all sets in the lattice up to the solution level and the mapping between levels. Unfortunately, this results in an exponential slowdown compared to the quantum implementation and severely limits the feasible size of these classical simulations. Moreover, determining the expected behavior of the random phase method requires repeating the search a number of times on each problem (10 tries in the experiments reported here). This further limits the feasible problem size.

As a simple check on the numerical errors of the calculation, we recorded the total normalization in all sets at the solution level. With double precision calculations on a Sun Sparc10, for the experiments reported here typically the norm was 1 to within a few times $10^{-11}$. As an indication of the execution time with unoptimized C++ code, a single trial for a problem with $N = 14$ and 16, with $L = N/2$, required about 70 and 1000 seconds, respectively. This uses a direct evaluation of the map from one level to the next as given by Eq. 21. A substantial reduction in compute time is possible by exploiting the simple structure of this mapping to give a recursive evaluation[10]. Some additional improvement is possible by exploiting the fact that all amplitudes are real when using the method that inverts phases of nogoods. This reduced the execution time to about 1 and 6 seconds per trial for $N$ of 14 and 16, respectively.

## 4.1 Unstructured Problems

To examine the typical behavior of this quantum search algorithm with respect to problem structure, we need a suitable class of problems. This is particularly important for average case analyses since one could inadvertently select a class of search problems dominated by easy cases. Fortunately the observed concentration of hard cases near phase transitions provides a method to generate hard test cases.

The phase transition behavior has been seen in a variety of search problem classes. Here we select a particularly simple class of problems by supposing the constraints specify nogoods randomly at level 2 in the lattice. This corresponds to binary constraint satisfaction problems (Prosser, 1996; Smith & Dyer, 1996), but ignores the detailed structure of the nogoods imposed by the requirement that variables have a unique assignment. By ignoring this additional structure, we are able to test a wider range of the number of specified nogoods for the problems than would be the case by considering only constraint satisfaction problems. This lack of additional structure is also likely to make the asymptotic behavior more readily apparent at the small problem sizes that are feasible with a classical simulation.

Furthermore, since the quantum search algorithm is appropriate only for soluble problems, we restrict attention to random problems with a solution. These could be obtained by randomly generating problems and rejecting any that have no solution (as determined using a complete classical search method). However, for overconstrained problems the soluble ones become quite rare and difficult to find by this method. Instead, we generate problems with a prespecified solution. That is, when randomly selecting nogoods to add to a problem, we do not pick any nogoods that are subsets of a prespecified solution set. This always produces problems with at least one solution. Although these problems tend to be a bit easier than randomly selected soluble problems, they nevertheless exhibit the same concentration of hard problems and at about the same location as general random problems (Cheeseman et al., 1991; Williams & Hogg, 1994). The quantum search is started at level 2 in the lattice.

---

10. I thank S. Vavasis for suggesting this improvement in the classical simulation of the algorithm.

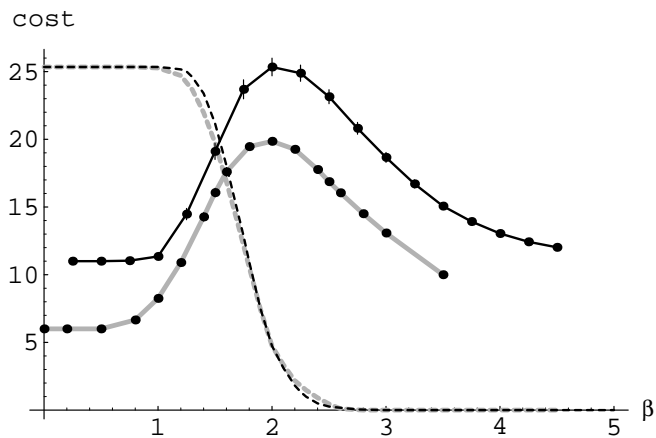

Figure 3: The solid curves show the classical backtrack search cost for randomly generated problems with a prespecified solution as a function of $\beta = m/N$ for $N = 10$ (gray) and 20 (black) and $L = N/2$. Here $m$ is the number of nogoods selected at level 2 of the search lattice. The cost is the average number of backtrack steps, starting from the empty set, required to find the first solution to the problem, averaged over 1000 problems. The error bars indicate the standard deviation of this estimate of the average value, and in most cases are smaller than the size of the plotted points. For comparison, the dashed curves show the probability for having a solution in randomly generated problems with the specified $\beta$ value, ranging from 1 at the left to 0 at the right.

### 4.1.1 Theory

For this class of problems, the phase transition behavior is illustrated in Fig. 3. Specifically, this shows the cost to solve the problem with a simple chronological backtrack search. The cost is given in terms of the number of search nodes considered until a solution is found. The minimum cost, for a search that proceeds directly to a solution with no backtrack is $L + 1$. The parameter distinguishing underconstrained from overconstrained problems is the ratio $\beta$ of the number of nogoods $m$ at level 2 given by the constraints to the number of items $N$.

Even for these relatively small problems, a peak in the average search cost is evident. Moreover, this peak is near the transition region where random problems[11] change from mostly soluble to mostly insoluble. A simple, but approximate, theoretical value for the location of the transition is given by the point where the expected number of solutions is equal to one (Smith & Dyer, 1996; Williams & Hogg, 1994). Applying this to the class of problems considered here is straightforward. Specifically, there are $N_L$ complete sets for the problem, as given by Eq. 1. A particular set $s$ of size $L$ will be good, i.e., a solution, only if none of the nogoods selected for the problem are a subset of $s$. Hence the probability it

---

11. That is, problems generated by random selection of nogoods without regard for whether they have a solution.

will be a solution is given by

$$\rho_L = \frac{\left( \binom{\binom{N}{2} - \binom{L}{2}}{m} \right)}{\left( \binom{\binom{N}{2}}{m} \right)} \tag{25}$$

because there are $\binom{N}{2}$ sets of size 2 from which to choose the $m$ nogoods specified directly by the constraints. The average number of solutions is then just $N_{soln} = N_L \rho_L$. If we set $m = \beta N$ and $L = N/b$, for large $N$ this becomes

$$\ln N_{soln} \sim N \left( h\left(\frac{1}{b}\right) + \beta \ln \left(1 - \frac{1}{b^2}\right) \right) \tag{26}$$

where $h(x) \equiv -x \ln x - (1-x) \ln (1-x)$. The predicted transition point[12] is then given by

$$\beta_{crit} = \frac{h(1/b)}{-\ln (1 - 1/b^2)} \tag{27}$$

which is $\beta_{crit} = 2.41$ for the case considered here (i.e., $b = 2$). This closely matches the location of the peak in the search cost for problems with prespecified solution, as shown in Fig. 3, but is about 20% larger than the location of the step in solubility[13]. Furthermore, the theory predicts there is a regime of polynomial average cost for sufficiently few constraints (Hogg & Williams, 1994). This is determined by the condition that the expected number of goods at each level in the lattice is monotonically increasing. Repeating the above argument for smaller levels in the lattice, we find that this condition holds up to

$$\beta_{poly} = \frac{b^2 - 1}{2b} \ln(b - 1) \tag{28}$$

which is $\beta_{poly} = 0$ for $b = 2$.

While these estimates are only approximate, they do indicate that the class of random soluble problems defined here behaves qualitatively and quantitatively the same with respect to the transition behavior as a variety of other, perhaps more realistic, problem classes. This close correspondence with the theory (derived for the limit of large problems), suggests that we are observing the correct transition behavior even with these relatively small problems. Moreover the above approximate theoretical argument suggests that the average cost of general classical search methods scales exponentially with the size of the problem over the full range of $\beta > 0$. Thus this provides a good test case for the average behavior of the quantum algorithm. As a final observation, it is important to obtain a sufficient number of samples, especially near the transition region. This is because there is considerable variation in problems near the transition, specifically a highly skewed distribution in the number of solutions. In this region, most problems have few solutions but a few have extremely many: enough in fact to give a substantial contribution to the average number of solutions even though such problems are quite rare.

---

12. This differs slightly from the results for problems with more specified structure on the nogoods, such as explicitly removing the necessary nogoods from consideration (Smith & Dyer, 1996; Williams & Hogg, 1994).

13. This is a particularly large error for this theory: it does better for problems with larger constraints or more allowed values per variable.

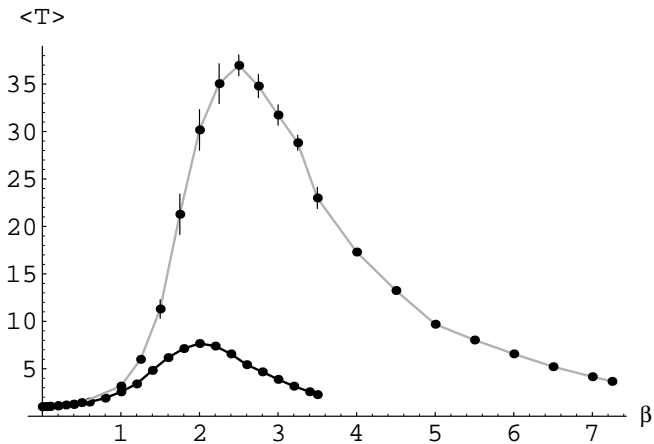

Figure 4: Expected number of trials $\langle T \rangle$ to find a solution vs. $\beta$ for random problems with prespecified solution with binary constraints, using random phases for nogoods. The solid curve is for $N = 10$, with 100 samples per point. The gray curve is for $N = 20$ with 10 samples per point (but additional samples were used around the peak). The error bars indicate the standard error in the estimate of $\langle T \rangle$.

### 4.1.2 Phase Transition

To see how problem structure affects this search algorithm, we evaluate $p_{soln}$, the probability to find a solution for problems with different structures, ranging from underconstrained to overconstrained. Low values for this probability indicate relatively harder problems. The expected number of repetitions of the search required to find a solution is then given by $T = 1/p_{soln}$. The results are shown in Figs. 4 and 5 for different ways of introducing phases for nogood sets. We see the general easy-hard-easy pattern in both cases. Another common feature of phase transitions is an increased variance around the transition region. The quantum search has this property as well, as shown in Fig. 6.

### 4.1.3 Scaling

An important question in the behavior of this search method is how its average performance scales with problem size. To examine this question, we consider the scaling with fixed $\beta$. This is shown in Figs. 7 and 8 for algorithms using random and inverted phases for nogoods, respectively. To help identify the likely scaling, we show the same results on both a log plot (where straight lines correspond to exponential scaling) and a log-log plot (where straight lines correspond to power-law or polynomial scaling).

It is difficult to make definite conclusions from these results for two reasons. First, the variation in behavior of different problems gives a statistical uncertainty to the estimates of the average values, particularly for the larger sizes where fewer samples are available. The standard errors in the estimates of the averages are indicated by the error bars in the figures (though in most cases, the errors are smaller than the size of the plotted points). Second, the scaling behavior could change as larger cases are considered. With these caveats in mind, the figures suggest that $p_{soln}$ remains nearly constant for underconstrained problems, even though the fraction of complete sets that are solutions is decreasing exponentially. This behavior is also seen in the overlap of the curves for small $\beta$ in Figs. 4 and 5. For problems with more constraints, $p_{soln}$ appears to decrease polynomially with the size of the problem, i.e., the curves are closer to linear in the log-log plots than in the log plots. This in confirmed quantitatively by making a least squares fit to the values and seeing that the residuals of the fit to a power-law are smaller than those for an exponential fit. An interesting observation in comparing the two phase choices is that the scaling is qualitatively similar, even though the phase inversion method performs better. This suggests the detailed values of the phase choices are not critical to the scaling behavior, and in particular high precision evaluation of the phases is not required. Finally we should note that this illustration of the average scaling leaves open the behavior for the worst case instances.

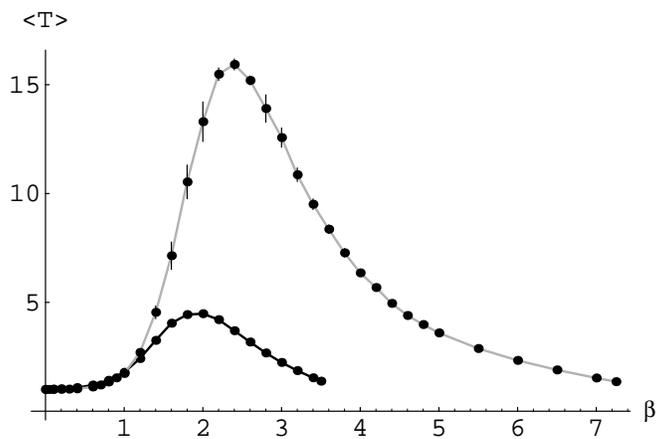

Figure 5: Expected number of trials $\langle T \rangle$ to find a solution vs. $\beta$ for random problems with prespecified solution with binary constraints, using inverted phases for nogoods. The solid curve is for $N = 10$, with 1000 samples per point. The gray curve is for $N = 20$ with 100 samples per point (but additional samples were used around the peak). The error bars indicate the standard error in the estimate of $\langle T \rangle$.

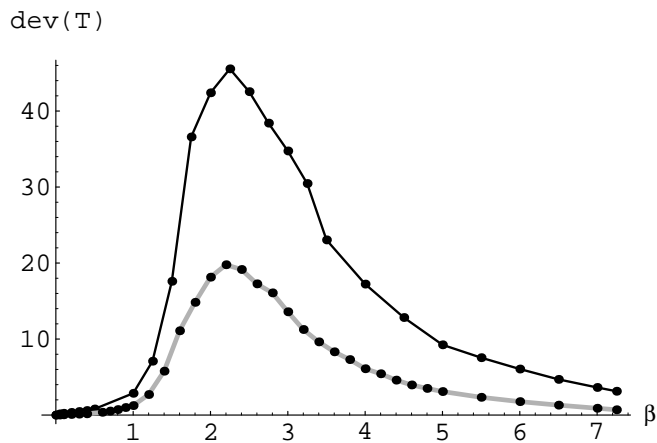

Figure 6: Standard deviation in the number of trials to find a solution for $N = 20$ as a function of $\beta$. The black curve is for random phases assigned to nogoods, and the gray one for inverting phases.

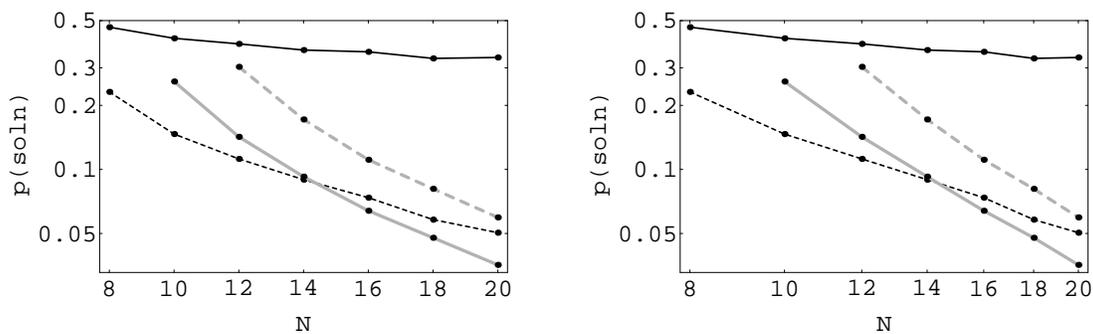

Figure 7: Scaling of the probability to find a solution using the random phase method, for $\beta$ of 1 (solid), 2 (dashed), 3 (gray) and 4 (dashed gray). This is shown on log and log-log scales (left and right plots, respectively).

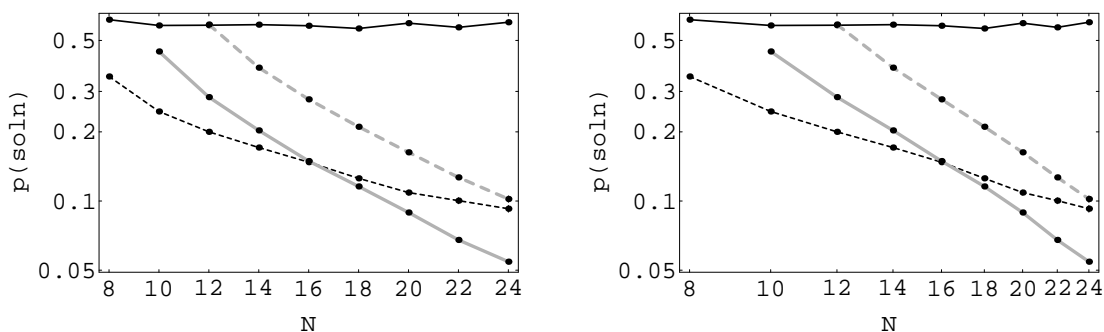

Figure 8: Scaling of the probability to find a solution using the phase inversion method, for $\beta$ of 1 (solid), 2 (dashed), 3 (gray) and 4 (dashed gray). This is shown on log and log-log scales (left and right plots, respectively).

For the underconstrained cases in Figs. 7 and 8 there is a small additional difference between cases with an even and odd number of variables. This is due to oscillations in the amplitude in goods at each level of the lattice, and is discussed more fully in the context of SAT problems below.

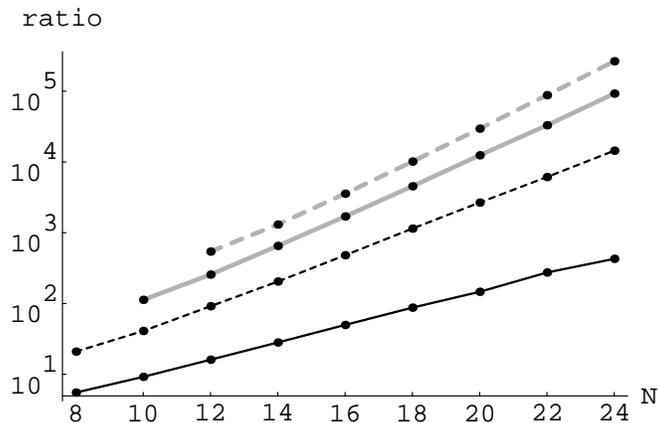

Figure 9: Scaling of the ratio of the probability to find a solution using the quantum algorithm to the probability to find a solution by random selection at the solution level, using the phase inversion method, for $\beta$ of 1 (solid), 2 (dashed), 3 (gray) and 4 (dashed gray). The curves are close to linear on this log scale indicating exponential improvement over the direct selection from among complete sets, with a higher enhancement for problems with more constraints.

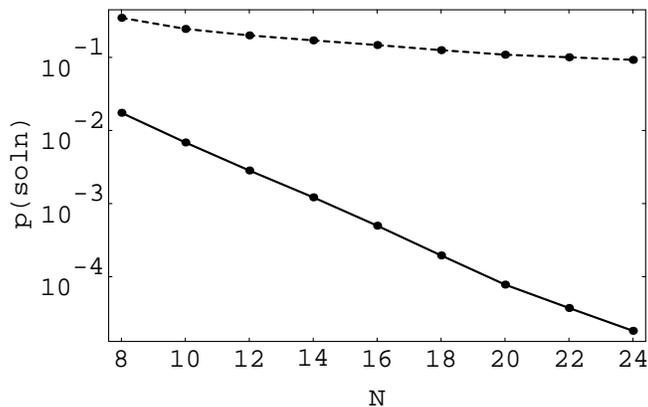

Figure 10: Comparison of scaling of probability to find a solution with the quantum algorithm using the phase inversion method (dashed curve) and by random selection at the solution level (solid curve) for $\beta = 2$.

Another scaling comparison is to see how much this algorithm enhances the probability to find a solution beyond the simple quantum algorithm of evaluating all the complete sets and then making a measurement. As shown in Fig. 9, this quantum algorithm appears to give an exponential improvement in the concentration of amplitude into solutions. A more explicit view of this difference in behavior is shown in Fig. 10 for $\beta = 2$. In this figure, the dashed curve shows the behavior of $p_{soln}$ for the phase inversion method, and is identical to the $\beta = 2$ curve of Fig. 8.

117

## 4.2 Random 3SAT

These experiments leave open the question of how additional problem structure might affect the scaling behaviors. While the universality of the phase transition behavior in other search methods suggests that the average behavior of this algorithm will also be the same for a wide range of problems, it is useful to check this empirically. To this end the algorithm was applied to the satisfiability (SAT) problem. This constraint satisfaction problem consists of a propositional formula with $n$ variables and the requirement to find an assignment (true or false) to each variable that makes the formula true. Thus there are $b = 2$ assignments for each variable and $N = 2n$ possible variable-value pairs. We consider the well-studied NP-complete 3SAT problem where the formula is a conjunction of $c$ clauses, each of which is a disjunction of 3 (possibly negated) variables.

The SAT problem is readily represented by nogoods in the lattice of sets (Williams & Hogg, 1994). As described in Sec. 2.2, there will be $n$ necessary nogoods, each of size 2. In addition, each distinct clause in the proposition gives a single nogood of size 3. This case is thus of additional interest in having specified nogoods of two sizes. For evaluating the quantum algorithm, we start at level 3 in the lattice. Thus the smallest case for which the phase choices will influence the result is for $n = 5$.

We generate random problems with a given number of clauses by selecting that number of different nogoods of size 3 from among those sets not already excluded by the necessary nogoods[14]. For random 3SAT, the hard problems are concentrated near the transition (Mitchell et al., 1992) at $c = 4.2n$. Finally, from among these randomly generated problems, we use only those that do in fact have a solution[15]. Using randomly selected soluble problems results in somewhat harder problems than using a prespecified solution. Like other studies that need to examine many goods and nogoods in the lattice (Schrag & Crawford, 1996), these results are restricted to much smaller problems than in most studies of random SAT. Consequently, the transition region is rather spread out. Furthermore, the additional structure of the necessary nogoods and the larger size of the constraints, compared with the previous class of problems, makes it more likely that larger problems will be required to see the asymptotic scaling behavior. However, at least some asymptotic behaviors have been observed (Crawford & Auton, 1993) to persist quite accurately even for problems as small as $n = 3$, so some indication of the scaling behavior is not out of the question for the small problems considered here.

### 4.2.1 Phase Transition

The behavior of the algorithm as a function of the ratio of clauses to variables is shown in Fig. 11 using the phase inversion method. This shows the phase transition behavior. Comparing to Fig. 5, this also shows the class of random 3SAT problems is harder, on average, for the quantum algorithm than the class of unstructured problems.

---

14. This differs slightly from other studies of random 3SAT in not allowing duplicate clauses in the propositional formula.

15. For the values of $c/n$ and small problems examined here, there are enough soluble instances randomly generated that there is no need to rely on a prespecified solution to efficiently find soluble test problems.

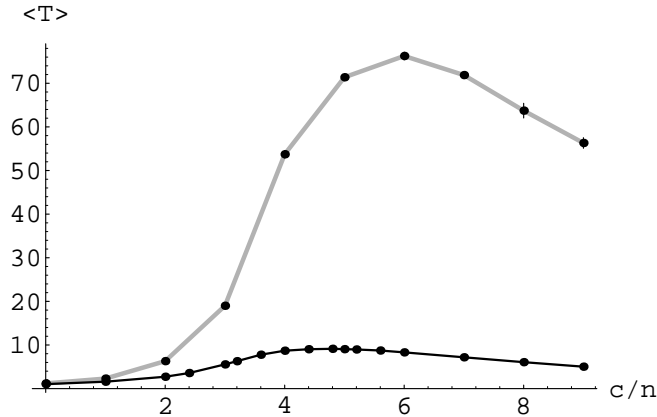

Figure 11: Average number of tries to find a solution with the quantum search algorithm for random 3SAT as a function of $c/n$, using the phase inversion method. The curves correspond to $n = 5$ (black) and $n = 10$ (gray).

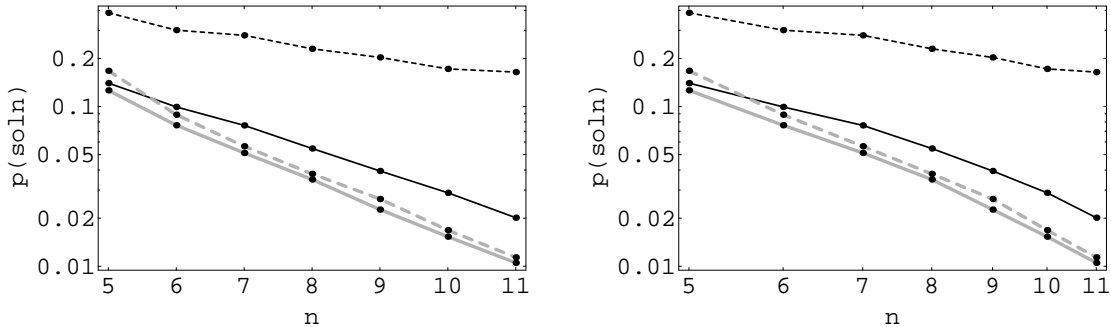

Figure 12: Scaling of the probability to find a solution, using the phase inversion method, as a function of the number of variables for random 3SAT problems. The curves correspond to different clause to variable ratios: 2 (dashed), 4 (solid), 6 (gray) and 8 (gray, dashed). This is shown on log and log-log scales (left and right plots, respectively).

### 4.2.2 SCALING

The scaling of the probability to find a solution is shown in Fig. 12 using the phase inversion method. More limited experiments with the random phase method showed the same behavior as seen with the unstructured class of problems: somewhat worse performance but similar scaling behavior. The results here are less clear-cut than those of Fig. 8. For $c/n = 2$ the results are consistent with either polynomial or exponential scaling. For problems with more constraints, exponential scaling is a somewhat better fit.

In addition to the general scaling trend, there is also a noticeable difference in behavior between cases with an even and odd number of variables. This is due to the behavior of the amplitude at each step in the lattice. Instead of a monotonic decrease in the concentration of amplitude into goods, there is an oscillatory behavior in which amplitude alternates between dispersing and being focused into goods at different levels. An extreme example of this behavior is shown in Fig. 13 for 3SAT problems with no constraints, i.e., $c = 0$. Specifically, at level $i$ this shows $\sum_s \left| \psi_s^{(i)} \right|^2$ where the sum is over all sets $s$ at level $i$ in the lattice that are consistent, which, for these problems with no constraints, are all assignments to $i$ variables. This is the probability that a good would be found if the algorithm were terminated at level $i$ and gives an indication of how well the algorithm concentrates amplitude among consistent states. In this case, the expanded search space of the quantum algorithm results in slightly worse performance than random selection from among complete assignments (all of which are solutions in this case). Each search starts with all amplitude in goods at level 3. Then the total probability in goods alternately decreases and increases as the map proceeds up to the solution level. Cases with an even number of variables (the black curves in the figure) end on a step that decreases the probability in goods, resulting in relatively lower performance compared to the odd variable cases (gray curves). Although this might suggest an improvement for the even $n$ cases by starting in level 2 rather than level 3, in fact this turns out not to be the case: starting in level 2 gives essentially the same behavior for the upper levels as starting the search from level 3 of the lattice due to one oscillation at intermediate levels that takes 2 steps to complete. Increasing the value of $c/n$, i.e., examining SAT problems with constraints, reduces the extent of the oscillations, particularly in higher levels of the lattice, and eventually results in monotonic decrease in probability as the search moves up the lattice. Nevertheless, for problems with a few constraints the existence of these oscillations gives rise to the observed difference in behavior between cases with an even and odd number of variables. These oscillations are also seen for underconstrained cases of unstructured problems in Figs. 7 and 8.

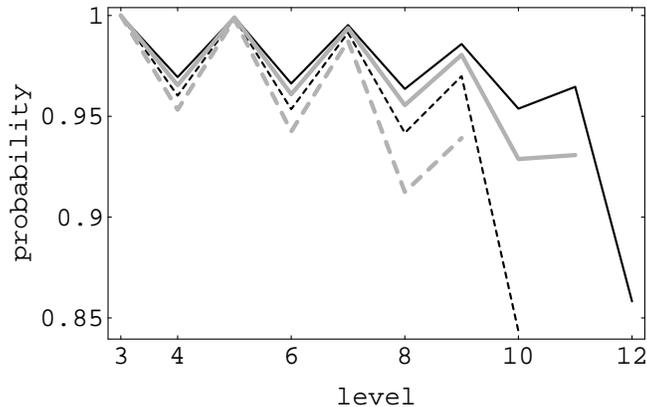

Figure 13: Probability in goods (i.e., consistent sets) as a function of level in the lattice for 3SAT problems with no constraints. This shows the behavior for $n$ equal to 9 (gray dashed), 10 (black dashed), 11 (gray) and 12 (black). For each problem, the final probability at level $n$ is the probability a solution is obtained with the quantum algorithm.

While Fig. 13 shows that the oscillatory behavior decreases for larger problems, it also suggests there may be more appropriate choices of the phases. Specifically, it may be possible to obtain a greater concentration of amplitude into solutions by allowing more dispersion into nogoods at intermediate levels of the lattice or using an initial condition with some amplitude in nogoods. If so, this would represent a new policy for selecting the phases that takes into account the problem-independent structure of the necessary nogoods. This would be somewhat analogous to focusing light with a lens: paths in many directions are modified by the lens to cause a convergence to a single point.

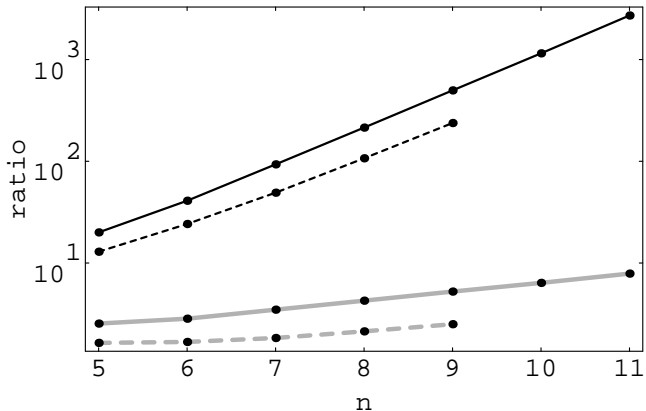

Figure 14: Scaling of the ratio of the probability to find a solution using the quantum algorithm to the probability to find a solution by random selection at the solution level as a function of the number of variables for random 3SAT problems with clause to variable ratio equal to 4. The solid and dashed curves correspond to using the phase inversion and random phase methods, respectively. The black curves compare to random selection among complete sets, while the gray compare to selection only from among complete assignments. The curves are close to linear on this log scale indicating exponential improvement over the direct selection from among complete sets.

More definite results are obtained for the improvement over random selection. Specifically, Fig. 14 shows an exponential improvement for both the phase inversion and random phase methods, corresponding to the behavior for unstructured problems in Fig. 9. Similar improvement is seen for other values of $c/n$ as well: as in Fig. 9 the more highly constrained problems give larger improvements. A more stringent comparison is with random selection from among complete assignments (i.e., each variable given a single value) rather than from among all complete sets of variable-value pairs. This is also shown in Fig. 14, appearing to grow exponentially as well. This is particularly significant because the quantum algorithm uses a larger search space containing the necessary nogoods. Another view of this comparison is given in Fig. 15, showing the probabilities to find a solution with the quantum search and random selection from among complete assignments. We conclude from these results that the additional structure of necessary nogoods and constraints of different sizes is qualitatively similar to that for unstructured random problems but a detailed comparison of the scaling behaviors requires examining larger problem sizes.

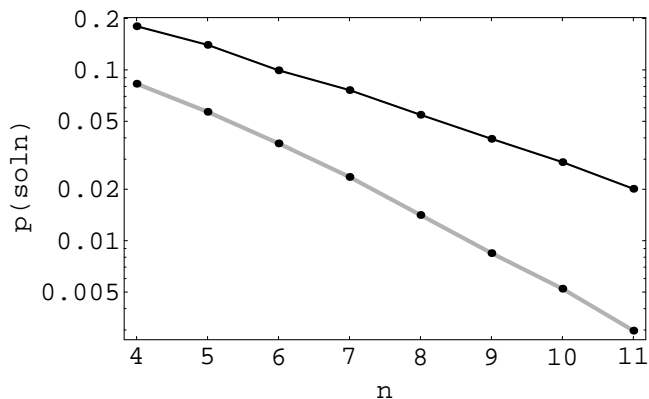

Figure 15: Comparison of scaling of probability to find a solution with the quantum algorithm using the phase inversion method (solid curve) and by random selection from among complete assignments (gray curve) for $c/n = 4$.

## 5. Discussion

In summary, we have introduced a quantum search algorithm and evaluated its average behavior on a range of small problems. It appears to increase the amplitude into solution states exponentially compared to evaluating and measuring a quantum superposition of potential solutions directly. Moreover, this method exhibits the same transition behavior, with its associated concentration of hard problems, as seen with many classical search methods. It thus extends the range of methods to which this phenomenon applies. More importantly, this indicates the algorithm is effectively exploiting the same structure of search problems as, say, classical backtrack methods, to prune unproductive search directions. It is thus a major improvement over the simple applications of quantum computing to search problems that behave essentially the same as classical generate-and-test, a method that completely ignores the possibility of pruning and hence doesn't exhibit the phase transition.

The transition behavior is readily understood because problems near the transition point have many large partial goods that do not lead to solutions (Williams & Hogg, 1994). Thus there will be a relatively high proportion of paths through the lattice that appear good for quite a while but eventually give deadends. A choice of phases based on detecting nogoods will not be able to work on these paths until near the solution level and hence give less chance to cancel out or move amplitude to those paths that do in fact lead to solutions. Hence problems with many large partial goods are likely to prove relatively difficult for any quantum algorithms that operate by distinguishing goods from nogoods of various sizes.

There remain many open questions. In the algorithm, the division between a problem–independent mapping through the lattice and a simple problem-specific adjustment to phases allows for a range of policies for selecting the phases. It would be useful to understand the effect of different policies in the hope of improving the concentration of amplitude into solutions. For example, the use of phases has two distinct jobs: first, to keep amplitude moving up along good sets rather than diffusing out to nogoods, and second, when a deadend is reached (i.e., a good set that has no good supersets) to send the amplitude at this deadend to a promising region of the search space, possibly very far from where the deadend occurred. These goals, of keeping amplitude concentrated on the one hand and sending it away on the other, are to some extent contradictory. Thus it may prove worthwhile to consider different phase choice policies for these two situations. Furthermore, the mapping through the lattice is motivated by classical methods that incrementally build solutions by moving from sets to supersets in the lattice. Instead of using unitary maps at each step that are as close as possible to this classical behavior, other approaches could allow more significant spreading of the amplitude at intermediate levels in the lattice and only concentrate it into solutions in the last few steps. It may prove fruitful to consider another type of mapping based on local repair methods moving among neighbors of complete sets. In this case, sets are evaluated based on the number of constraints they violate so an appropriate phase selection policy could depend on this number, rather than just whether the set is inconsistent or not. These possibilities may also suggest new probabilistic classical algorithms that might be competitive with existing heuristic search methods.

As a new example of a search method exhibiting the transition behavior, this work raises the same issues as prior studies of this phenomenon. For instance, to what extent does this behavior apply to more realistic classes of problems, such as those with clustering inherent in situations involving localized interactions (Hogg, 1996). This will be difficult to check empirically due to the limitation to small problems that are feasible for a classical simulation of this algorithm. However the observation that this behavior persists for many classes of problems with other search methods suggests it will be widely applicable. It is also of interest to see if other phase transition phenomena appear in these quantum search algorithms, such as observed in optimization searches (Cheeseman et al., 1991; Pemberton & Zhang, 1996; Zhang & Korf, 1996; Gent & Walsh, 1995). There may also be transitions unique to quantum algorithms, for example in the required coherence time or sensitivity to environmental noise.

For the specific instances of the algorithm presented here, there are also some remaining issues. An important one is the cost of the mapping from one level to the next in terms of more basic operations that might be realized in hardware, although the simple structure of the matrices involved suggest this should not be too costly. The scaling behavior of the algorithm for larger cases is also of interest, which can perhaps be approached by examining the asymptotic nature of the matrix coefficients of Eqs. 17 and 19.

An important practical question is the physical implementation of quantum computers in general (Barenco et al., 1995; Sleator & Weinfurter, 1995; Cirac & Zoller, 1995), and the requirements imposed by the algorithm described here. Any implementation of a quantum computer will need to deal with two important difficulties (Landauer, 1994). First, there will be defects in the construction of the device. Thus even if an ideal design exactly produces the desired mapping, occasional manufacturing defects and environmental noise will introduce errors. We thus need to understand the sensitivity of the algorithm's behavior to errors in the mappings. Here the main difficulty is likely to be in the problem-independent mapping from one level of the lattice to the next, since the choice of phases in the problem-specific part doesn't require high precision. In this context we should note that standard error correction methods cannot be used with quantum computers in light of the requirement that all operations are reversible. We also need to address the extent to which such errors can be minimized in the first place, thus placing less severe requirements on the algorithm. Particularly relevant in this respect is the possibility of drastically reducing defects in manufactured devices by atomically precise control of the hardware (Drexler, 1992; Eigler & Schweizer, 1990; Muller et al., 1995; Shen, Wang, Abeln, Tucker, Lyding, Avouris, & Walkup, 1995). There are also uniquely quantum mechanical approaches to controlling errors (Berthiaume, Deutsch, & Jozsa, 1994) based on partial measurements of the state. This work could substantially extend the range of ideal quantum algorithms that will be possible to implement.

The second major difficulty with constructing quantum computers is maintaining coherence of the superposition of states long enough to complete the computation. Environmental noise gradually couples to the state of the device, reducing the coherence and eventually limiting the time over which a superposition can perform useful computations (Unruh, 1995; Chuang, Laflamme, Shor, & Zurek, 1995). In effect, the coupling to the environment can be viewed as performing a measurement on the quantum system, destroying the superposition of states. This problem is particularly severe for proposed universal quantum computers that need to maintain superpositions for arbitrarily long times. In the method presented here, the number of steps is known in advance and could be implemented as a special purpose search device (for problems of a given size) rather than as a program running on a universal computer. Thus a given achievable coherence time would translate into a limit on feasible problem size. To the extent that this limit can be made larger than feasible for alternative classical search methods, the quantum search could be useful.

The open question of greatest theoretical interest is whether this algorithm or simple variants of it can concentrate amplitude into solutions sufficiently to give a polynomial, rather than exponential, decrease in the probability to find a solution of *any* NP search problem with small constraints. This is especially interesting since this class of problems includes many well-studied NP-complete problems such as graph coloring and propositional satisfiability. Even if this is not so in the worst case, it may be so on average for some classes of otherwise difficult real-world problems. While it is by no means clear to what extent quantum coherence provides more powerful computational behavior than classical machines, a recent proposal for rapid factoring (Shor, 1994) is an encouraging indication of its capabilities.

A more subtle question along these lines is how the average scaling behaves away from the transition region of hard problems. In particular, can such quantum algorithms expand the range of the polynomially scaling problems seen for highly underconstrained or overconstrained instances? If so, this would provide a class of problems of intermediate difficulty for which the quantum search is exponentially faster than classical methods, on average. This highlights the importance of broadening theoretical discussions of quantum algorithms to include typical or average behaviors in addition to worst case analyses. More generally, are there any differences in the phase transition behaviors or their location compared with the usual classical methods? These questions, involving the precise location of transition points, are not currently well understood even for classical search algorithms. Thus a comparison with the behavior of this quantum algorithm may help shed light on the nature of the various phase transitions that seem to be associated with the intrinsic structure of the search problems rather than with specific search algorithms.

## Acknowledgements

I thank John Gilbert, John Lamping and Steve Vavasis for their suggestions and comments on this work. I have also benefited from discussions with Peter Cheeseman, Scott Clearwater, Bernardo Huberman, Don Kimber, Colin Williams, Andrew Yao and Michael Youssefmir.